\documentclass{article}
\usepackage{spconf,amsmath,graphicx}
\usepackage{times}
\usepackage{xcolor}
\usepackage{epsfig}
\usepackage{amssymb}
\usepackage{bbm}
\usepackage[normalem]{ulem}
\useunder{\uline}{\ul}{}
\usepackage{multirow}
\newcommand{\etal}{\textit{et al.}}
\newcommand{\Lagr}{\mathcal{L}}

\title{Adversarial Pairwise Reverse Attention for Camera Performance Imbalance in Person Re-Identification: New Dataset and Metrics}
\name{Eugene P.W. Ang$^{\dagger \star}$ Shan Lin$^{\dagger \star}$ Rahul Ahuja$^{\ddagger}$ Nemath Ahmed$^{\mathsection}$ Alex C. Kot$^{\star}$\thanks{$^{\dagger}$Equal contribution.}}
\address{$^{\star}$Rapid-Rich Object Search (ROSE) Lab, Nanyang Technological University, Singapore\\
$^{\ddagger}$Computer Science Department, New York University, New York, USA\\
$^{\mathsection}$Department of Electrical Engineering, Indian Institute Technology Indore, Indore, India
}
\begin{document}
%
\maketitle
\begin{abstract}

Existing evaluation metrics for Person Re-Identification (Person ReID) models focus on system-wide performance. However, our studies reveal weaknesses due to the uneven data distributions among cameras and different camera properties that expose the ReID system to exploitation. In this work, we raise the long-ignored ReID problem of camera performance imbalance and collect a real-world privacy-aware dataset from 38 cameras to assist the study of the imbalance issue. We propose new metrics to quantify camera performance imbalance and further propose the \textbf{A}dversarial \textbf{P}airwise \textbf{R}everse \textbf{A}ttention (APRA) Module to guide the model towards learning camera invariant features with a novel pairwise attention inversion mechanism.

\end{abstract}
\section{Introduction}
\label{sec:intro}


Person re-identification, also known as Person ReID, deals with matching images or videos of the same person over a multiple-camera surveillance system. Although it enjoys rapid progress, issues remain during the transition to real-world deployment. For example, the standard metrics applied on a per-camera basis reveal that certain cameras in the network systematically under-perform their peers. Such vulnerabilities open the system up to exploitation by adversaries determined to evade it.

\begin{figure}[h]
\begin{center}
\begin{minipage}[t]{0.23\textwidth}
\includegraphics[clip,width=1\textwidth]{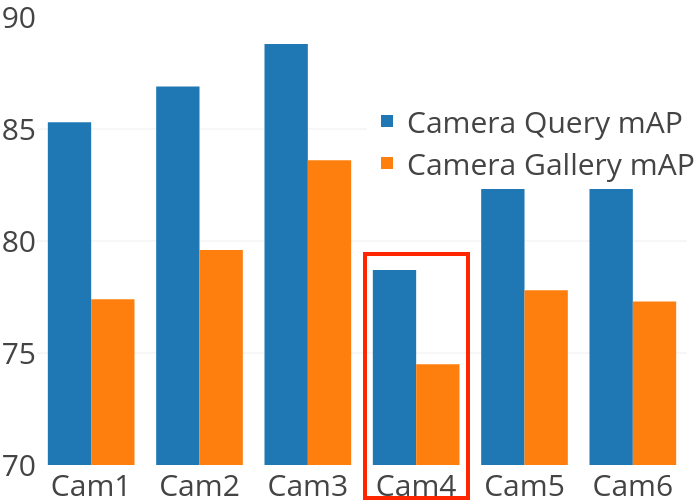}
\scriptsize \centering (a) Distribution of Camera Performance for Market-1501
\end{minipage}
\begin{minipage}[t]{0.23\textwidth}

\includegraphics[clip,width=1\textwidth]{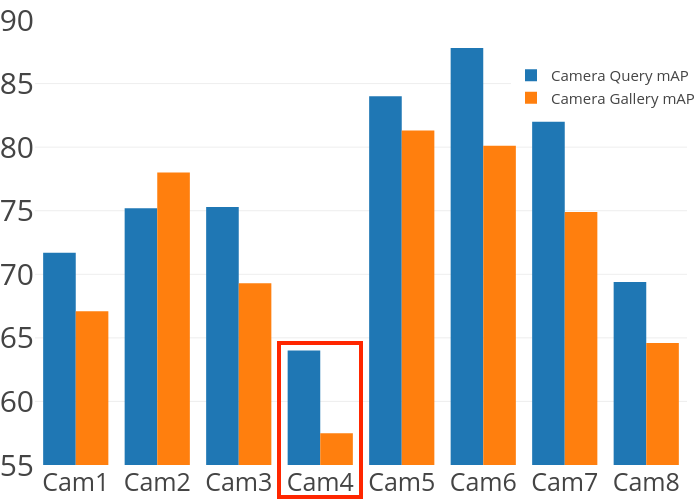}
\scriptsize \centering (b) Distribution of Camera Performance for DukeMTMC-reID
\end{minipage}
\end{center}
\vspace{-1em}
\caption{Camera Performance Imbalance Problem in Market-1501 and DukeMTMC-reID, outlined in red.}
\label{fig:imbalance}
\end{figure}

In order to quantify the performance imbalance of different cameras, we propose two new metrics for camera-specific evaluation: (1) camera query mean Average Precision (mAP) and (2) camera gallery mAP, which we shorten to \textbf{query-mAP} and \textbf{gallery-mAP} respectively. The query-mAP reports the performance when all query images are from one specific camera. On the other hand, the gallery-mAP measures the ease of retrieving positives from a particular camera. Figure \ref{fig:imbalance} (a) and (c) show the camera-level Person ReID performance of a baseline model trained on the Market-1501 \cite{Market-1501} and DukeMTMC-reID \cite{DukeMTMC-reID} datasets. Cameras 4 in both Market-1501 and DukeMTMC-reID datasets, outlined in red, lower the overall performance of the system.


Our proposed \textbf{A}dversarial \textbf{P}airwise \textbf{R}eversed \textbf{A}ttention (APRA) module leverages a novel pairwise attention inversion mechanism to disentangle camera features from identity features, utilizing adversarial gradient reversal to further suppress camera information. Adding our module to state-of-the-art methods better balances inter-camera performance on all four benchmarks studied in this paper.



To summarize, our contributions are as follows: (i) We identify and quantify inter-camera performance imbalance by formulating new per-camera metrics. (ii) We contribute a large-scale outdoor ReID dataset, NTU-Outdoor-38, boasting more cameras than other ReID datasets. (iii) We propose APRA, a module based on our novel paired attention inversion mechanism and apply it to state-of-the-art models to demonstrate better performance balance among cameras.

\section{Related Work}
\label{sec:related-work}

Modern Person ReID methods are mostly based on deep convolutional neural networks. Early deep learning based work usually formulated Person ReID as a verification problem and as a result solutions developed were based on Siamese architectures \cite{CUHK03, Huang2019Multi-PseudoRe-Identification, Lin2017End-to-EndRe-Identification}. In recent years, verification-driven approaches such as \cite{TriNet, Quadruplet} evolved from two-stream Siamese structures to triplet architectures to add robustness to the verification system. Another dominant approach is to formulate the Person ReID problem as a classification task. Zheng \etal \cite{IDE} first proposed the ID-discriminative embedding (IDE) to train an image to person ID mapping using models pre-trained on ImageNet \cite{ImageNet}. Sun \etal \cite{Sun2017SVDNetRetrieval} optimized the fully connected (FC) feature corresponding to person IDs with Singular Vector Decomposition (SVD). Many recent approaches \cite{Ensemble,MGN,StrongBaseline} consist of both classification loss and verification loss. The latest approaches, such as HA-CNN \cite{HA-CNN}, AANet \cite{AANet} and DuATM \cite{DuATM}, utilize attention mechanisms to further boost the Person Re-ID performance. 

Few studies have considered camera-level features and per-camera performance evaluation. Zhong \etal \cite{CamStyle} first discovered that image style variations caused by different cameras affect ReID retrieval results. They proposed a new data augmentation technique, CamStyle, which uses a CycleGAN \cite{CycleGAN} to generate new training images by combining appearance features from one camera and camera-style features from another camera. Zhang \etal \cite{SingleCamera} and Zhu \etal \cite{Zhu2019Intra-cameraBenchmark} train Person ReID models specialized on individual cameras. Unlike our study, the main motivation behind their work is to reduce ReID annotation labor by restricting the annotator's scope to a single camera. 
\section{NTU-Outdoor Dataset}
\subsection{Overview of Previous Datasets}
\vspace{-0.5em}
Compared to real-world video surveillance systems that keep track of hundreds of cameras, existing Person ReID datasets such as Market-1501 \cite{Market-1501}, DukeMTMC-reID \cite{DukeMTMC-reID} and MSMT17 \cite{MSMT17} contain a limited number (6-15) of cameras. This is primarily because annotation difficulty increases with the number of cameras, limiting its scalability. Additionally, most existing ReID datasets capture subjects without their awareness and consent. This lack of curation has raised privacy concerns from the public.

\vspace{-0.5em}
\subsection{Privacy-Aware Data Collection}
\vspace{-0.5em}
To address privacy/consent concerns, we apply a new privacy-aware data collection strategy during the collection of our new dataset. We developed a mobile application with functions for participants to declare consent, log their own appearance attributes, upload a reference picture of themselves and record their location over the duration of the data collection exercise to narrow the annotation window. We only annotate images belonging to consenting participants and discard images of other pedestrians.
\vspace{-0.5em}
\subsection{Dataset Characteristics}
\begin{figure}[h]
\centering
\includegraphics[width=0.48\textwidth]{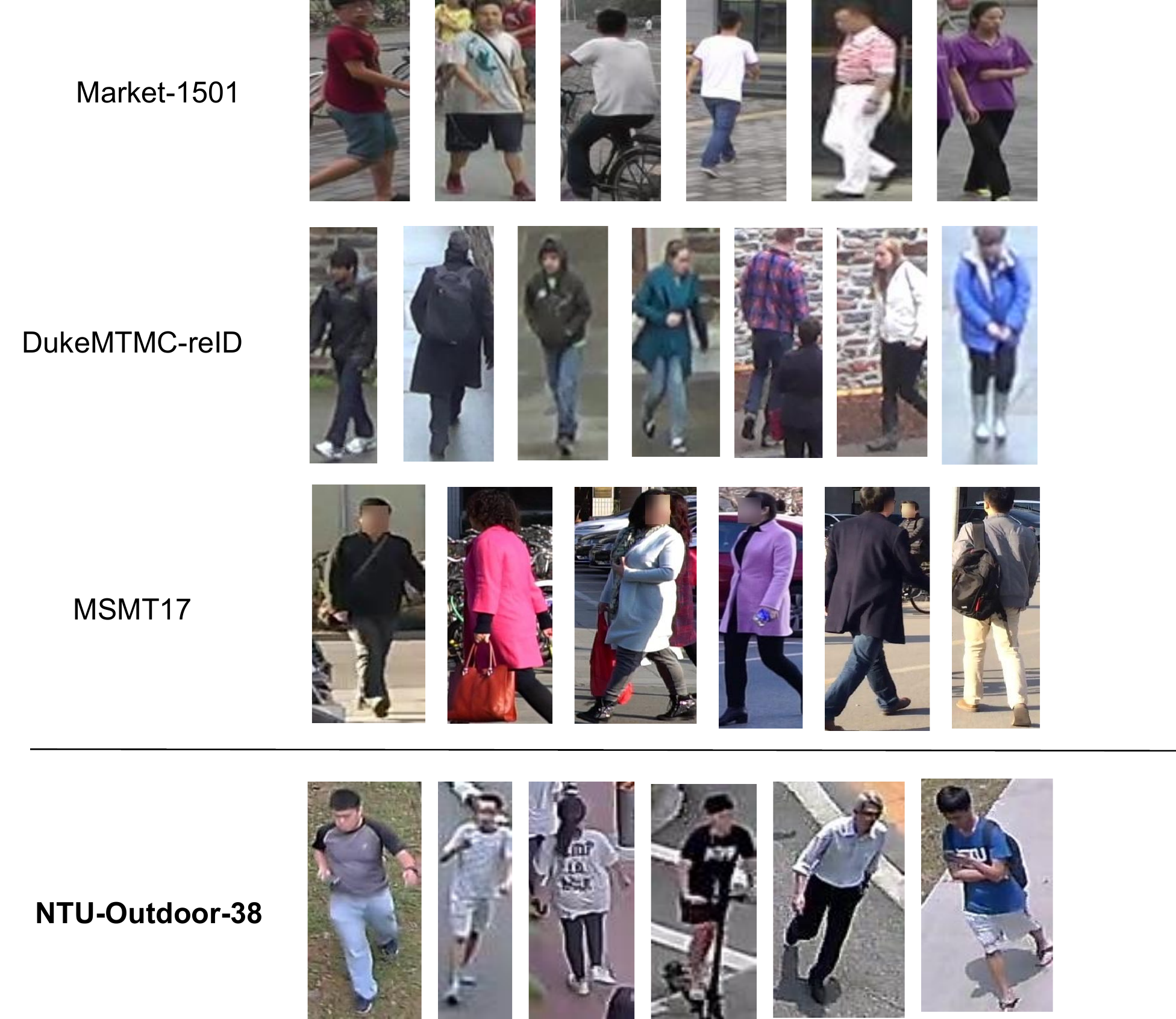}
\vspace{-2em}
\caption{Comparison of images in Market-1501, DukeMTMC-reID, MSMT17 and NTU-Outdoor-38.}
\label{fig:sample}
\end{figure}
\begin{table}[h]
\begin{center}
\resizebox{0.48\textwidth}{!}{%
\begin{tabular}{l|c|c|c|c}
\hline
Dataset      & NTU-Outdoor-38   & MSMT17      & DukeMTMC-reID & Market-1501 \\ \hline
\# Cameras   & \textbf{38}      & {\ul 15}          & 8             & 6           \\\hline
\# Images    & {\ul 48,347}     & \textbf{126,441}     & 36,411        & 32,668      \\\hline
\# Identites & 549              & \textbf{4,101}       & {\ul 1,812}   & 1,501       \\\hline
Privacy      & \textbf{Signed Agreement} & -           & -             & -           \\\hline
Camera Type     & \textbf{Surveillance}          & Normal & \textbf{Surveillance}           & Normal\\\hline
Detector     & YOLO V3          & Faster RCNN & DPM           & DPM         \\\hline
Attribute    & \textbf{40}      & -           & 23            & 30          \\ \hline
\end{tabular}}
\end{center}
\vspace{-1em}
\caption{Comparison between NTU-Outdoor-38 and other Person ReID datasets}
\label{tab:dataset}
\end{table}
\vspace{-0.5em}
Figure \ref{fig:sample} provides some sample images from Market-1501 \cite{Market-1501}, DukeMTMC-reID \cite{DukeMTMC-reID}, MSMT17 \cite{MSMT17}, and also our NTU-Outdoor-38 dataset. Market-1501 and MSMT17 used non-surveillance cameras, resulting in an unrealistic near-horizontal point of view of subjects. 

In the NTU-Outdoor-38 dataset, images are captured from actual surveillance cameras mounted on lamp-posts, better highlighting the viewing angles and imbalances inherent in real-world networked camera systems. It consists of outdoor scenes with large changes in viewpoint, illumination, and resolution that manifest even within individual wide-angle cameras. Our dataset spans 38 cameras, significantly more than other popular benchmarks. There are 549 appearance identities with signed privacy agreements for using their images for academic purposes. The NTU-Outdoor-38 dataset also comes with 40 additional binary attributes annotated by participants. Table \ref{tab:dataset} presents the characteristics of the NTU-Outdoor-38 dataset compared against Market-1501 \cite{Market-1501}, DukeMTMC-reID \cite{DukeMTMC-reID} and MSMT17 \cite{MSMT17}. NTU-Outdoor-38 captures the inter-camera performance imbalances present in real-life camera networks and serves as an excellent test-bed to further study this problem. To the best of our knowledge, it is the only publicly available Person ReID dataset collected from over 30 cameras.
\vspace{-0.5em}
\section{Camera-Level Evaluation Metrics}
\label{sec:cam-eval-metrics}


We present two new per-camera metrics to evaluate the performance of individual cameras in a ReID system. Query mAP (q-mAP) is a grouping of the query set by disjoint camera ids: $\text{mAP}_{c}^{q} = \frac{1}{|Q_c|} \sum_{q \in Q_c} \text{AP}(q, G)$, where $Q_c$ is the set of query images captured from Camera c, $G$ is the gallery set and AP is the average precision metric. 

Conversely, Gallery mAP (g-mAP) keeps only the positives from a chosen camera during retrieval, ignoring other positive candidates from other cameras. This helps us to evaluate the ease of retrieving positive gallery candidates from a chosen camera only. The g-mAP of Camera c is given as $\text{mAP}_{c}^{g} = \frac{1}{|Q_c|} \sum_{q \in Q_c} \text{AP}(q, G_{q})$, where $G_{q}$ is the gallery set excluding other positives not captured by Camera c and $Q_c$ is the set of queries that have at least one positive in the gallery from Camera c.

\section{Proposed Method}
\label{sec:method}

\begin{figure*}[h]
\centering
\includegraphics[width=1\textwidth]{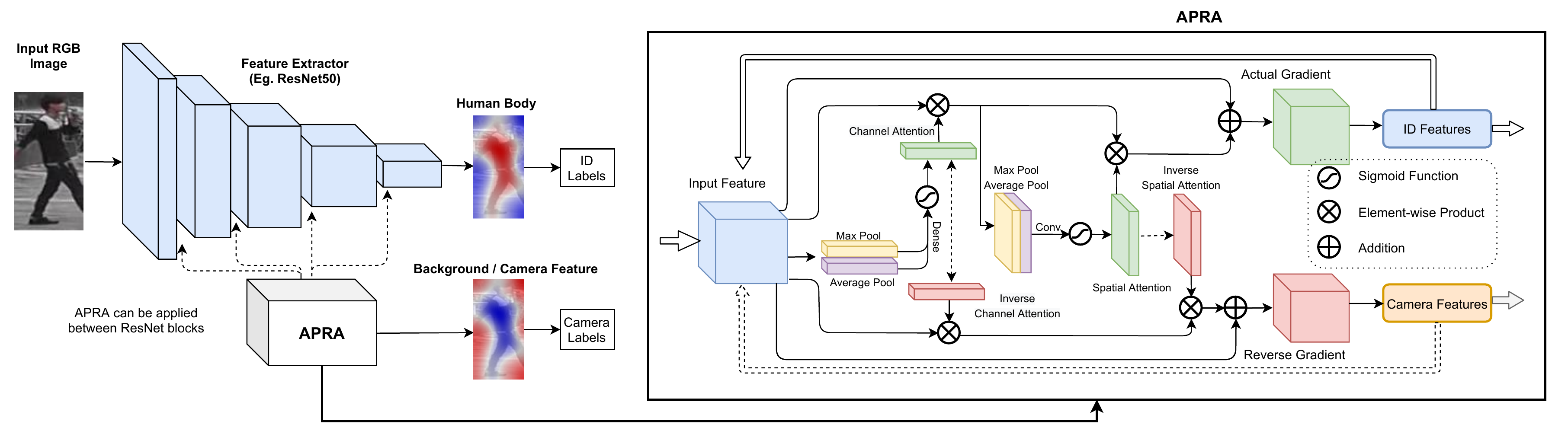}
\caption{\textbf{Left:} The APRA module is inserted in between convolutional blocks of a CNN, e.g. ResNet-50. \textbf{Right:} details of APRA. A dense layer derives the channel attention vector from pooled input features. A convolutional layer derives the corresponding spatial attention filter from channel-pooled features after channel modulation. The ID-branch uses the channel and spatial attention maps while the camera-branch uses the corresponding inverse maps.}
\vspace{-0.5em}
\label{fig:CAPRA}
\end{figure*}

\subsection{APRA Module}
Figure \ref{fig:CAPRA} illustrates the architecture of our APRA module. The APRA module is comprised of two key pieces: \textbf{Pairwise Reverse Attention} and \textbf{Adversarial Gradient Reversal}. First, it introduces a new camera classification branch. The input feature is dynamically divided into different branches via reversed pairwise attention, and we apply adversarial gradient reversal on the camera branch to encourage learning of camera-invariant ID features. Our modules can be placed in between layers of a standard deep convolutional neural network. Based on our experiments, placing the APRA module in early layers yields the best performance.



\vspace{-1em}
\subsubsection{Pairwise Reverse Attention}
\vspace{-0.5em}

Our APRA module, as shown in Figure \ref{fig:CAPRA} divides an input feature map between person and camera classification branches. This division is negotiated using attention mechanisms that operate on the channel and spatial dimensions of the feature map. Figure \ref{fig:CAPRA} visualizes one example of our learned in spatial attention maps. Our proposed module operates similar to a ``foreground'' and ``background'' separation where it naturally teaches the model to focus on the subject for id classification and on the surroundings for camera classification. Our APRA module performs the same disentanglement along the channel dimension, which is harder to visualize. 



Given a feature map $\boldsymbol{F} \in \mathbb{R}^{C \times H \times W}$, we follow \cite{CBAM} to derive a $\mathbb{R}^{C \times 1 \times 1}$ channel attention tensor $M_c$ and a $\mathbb{R}^{1 \times H \times W}$ spatial attention tensor $M_s$. We extend this technique by deriving an inverse channel attention $M_{c}' = 1 - M_c$ and an inverse spatial attention $M_{s}' = 1 - M_s$. The input to the person branch is given by $\boldsymbol{F}_{\rho} = [\: (\boldsymbol{F} \otimes M_{c})  \otimes M_{s} + \boldsymbol{F} \:]_{+}$, where $\otimes$ is element-wise multiplication with broadcasting. Conversely, the camera branch output is derived from the reverse attention maps $\boldsymbol{F}_{\kappa} = [ \: (\boldsymbol{F} \otimes M_{c}')  \otimes M_{s}' + \boldsymbol{F} \: ]_{+}$. Thus, our APRA module exploits reverse attention to generate a separate feature pathway for the camera classification branch.

Division of input features induces competition between both pathways, but can also be symbiotic: higher-level semantic features are suitable for person identification, whereas lower-level style-based features suit camera classification. By placing APRA among earlier layers of the base model (Figure \ref{fig:CAPRA}), we foster cooperation; low-level camera features are purged upstream, allowing the model to focus on learning id-relevant high-level features, reducing the tendency to over-fit and thus balancing performance between cameras.

\vspace{-1em}
\subsubsection{Adversarial Gradient Reversal}
\vspace{-0.5em}
Drawing inspiration from \cite{DANN}, during back-propagation we reverse the gradients by multiplying all gradients up to the camera output branch of the APRA module (Figure \ref{fig:CAPRA}) by a negative scalar. This adversarial setup encourages the model to learn camera-invariant features by reducing the tendency to over-fit to cameras.

\subsection{Combined Loss Function}
The first loss function we use is cross entropy, $\Lagr_{CE} = - \frac{1}{n} \sum_{i=1}^{n} \log( \: p_{\theta} (y_i | x_i) \: )$, where $x_i$ are the training images, $y_i$ are the corresponding (either person or camera identity) ground-truth labels, $n$ is the number of samples in the batch and $\theta$ are the model parameters. The second loss function is the triplet loss, $\Lagr_{triplet} = \frac{1}{|T|} \sum_{a,p,n \in T} [\: \delta_{\theta}(a,p) - \delta_{\theta}(a,n) + m \:]_{+}$, where $T$ is the set of triplets, $\delta$ is a distance metric and $m$ is a positive margin. The person identity loss is $\Lagr_{\rho} = \Lagr_{CE}^{Person} + \Lagr_{triplet}$. The camera identity loss is
$\Lagr_{\kappa} = \Lagr_{CE}^{Camera}$. Our proposed solution is a combination of both branch losses, $\Lagr = \Lagr_{\rho} + \lambda \Lagr_{\kappa}$. In all our experiments, the hyperparameter $\lambda$ = 0.01. Once the model is trained, we use the embeddings from the person branch.

\vspace{-0.5em}
\section{Experiments}
\subsection{Dataset and Setting}
\vspace{-0.5em}
We perform comparisons over three of the most popular benchmarks: Market-1501 \cite{Market-1501}, DukeMTMC-reID \cite{DukeMTMC-reID}, MSMT17 \cite{MSMT17} and our NTU-Outdoor-38 dataset, which we abbreviate to Market, Duke, MSMT and NTU-38 respectively. The statistics of each dataset are reported in Table \ref{tab:dataset}. Each dataset gets more challenging as the number of cameras increases. For system-wide performance evaluation, we adopt the widely used Rank-1 and mAP scores. For detailed camera level evaluation, we use our newly proposed query-mAP and gallery-mAP.


\subsection{Performance Evaluation}
\subsubsection{Camera Performance Evaluation}
\vspace{-0.5em}
We evaluate our method's improvement in camera-specific performance on Market, Duke, MSMT and NTU-38, using Q-mAP and G-mAP to denote query-mAP and gallery-mAP, respectively. As shown in Table \ref{tab:ComparisonHighLevel}, the weakest cameras reap the most improvements in all benchmarks, achieving more substantial gains compared to other cameras. Also, datasets that are more imbalanced by virtue of having more cameras exhibit a greater rebalancing effect, as shown by the significant improvements in MSMT (15 cameras) and NTU-38 (38 cameras). 

\vspace{-0.5em}
\begin{table}[h]
\begin{center}
\resizebox{0.48\textwidth}{!}{%
\begin{tabular}{l|c|cc|cc}
\hline
\multirow{2}{*}{Dataset} & \multirow{2}{*}{Method} & \multicolumn{2}{c|}{Weakest Camera} & \multicolumn{2}{c}{Average Camera} \\
 &  & Q-mAP & G-mAP & Q-mAP & G-mAP \\ \cline{1-6}
\multirow{3}{*}{Market} & BagTricks & 78.7 & 74.5 & 85.4 & 78.8  \\
 & +APRA & \textbf{80.5} & \textbf{76.2} & \textbf{86.6} & \textbf{80.6}\\
 & RGA & 81.6 & 77.2 & 87.5 & 81.5  \\
 & +APRA & \textbf{83.1} & \textbf{77.6} & \textbf{88.1} & \textbf{82.5}\\ \hline
\multirow{4}{*}{Duke} & BagTricks & 64.0 & 57.5 & 74.0 & 70.6  \\
 & +APRA & \textbf{68.6} & \textbf{61.0} & \textbf{77.7} & \textbf{74.8}  \\ 
 & RGA & 68.2 & 61.0 & 78.2 & 74.4  \\
 & +APRA & \textbf{72.0} & \textbf{63.0} & \textbf{78.6} & \textbf{74.8}  \\ \hline
\multirow{4}{*}{MSMT} & BagTricks & 17.4 & 20.0 & 48.4 & 35.4  \\
 & +APRA & \textbf{19.8} & \textbf{21.7} & \textbf{50.0} & \textbf{37.0}  \\
 & RGA & 28.0 & 25.5 & 51.8 & 39.0  \\
 & +APRA & \textbf{30.5} & \textbf{26.7} & \textbf{52.7} & \textbf{39.2} \\ \hline
\multirow{4}{*}{NTU-38} & BagTricks  & 19.6 & 8.9 & 33.3 & 17.1 \\
 & +APRA & \textbf{22.7} & \textbf{11.0} & \textbf{37.1} & \textbf{20.3}  \\ 
 & RGA  & 22.9 & 13.8 & 38.9 & 22.0 \\
 & +APRA & \textbf{27.2} & \textbf{15.1} & \textbf{39.9} & \textbf{23.0}\\\hline
\end{tabular}}
\end{center}
\vspace{-1em}
\caption{Average and bottleneck scores across datasets.}
\label{tab:ComparisonHighLevel}
\end{table}
\vspace{-2em}
\subsubsection{System-wide Performance Evaluation}
\vspace{-0.5em}
The second experiment evaluates our method's system-wide performance against state-of-the-art methods in Market and Duke. For comparison, we select three body-mask guided methods, MaskGuided \cite{MaskGuided} ,MaskReID \cite{MaskReID} and SPReID \cite{SPReID}, four attention-based methods, DuATM \cite{DuATM}, HA-CNN \cite{HA-CNN} Mancs \cite{Mancs} and AANet \cite{AANet}, and two GAN-based camera style transfer methods, Camstyle \cite{CamStyle} and PN-GAN \cite{PN-GAN}. To demonstrate the effectiveness of APRA, we used our baseline BagTricks \cite{StrongBaseline} and RGA \cite{RGA} with the APRA module added between their early feature layers. A detailed comparison of the state-of-the-art methods is shown in Table \ref{tab:SOTA}. Scores for RGA \cite{RGA} are lower than reported in the original paper as the official open source code was used to train their models and add APRA. Adding APRA to \cite{StrongBaseline} and \cite{RGA} improves their mAP scores, demonstrating that our APRA module does not sacrifice overall performance. 
\vspace{-0.5em}
\begin{table}[h]
\begin{center}
\resizebox{0.48\textwidth}{!}{%
\begin{tabular}{l|c|cc|cc}
\hline
\multicolumn{1}{c|}{\multirow{2}{*}{Type}} & \multirow{2}{*}{Method} & \multicolumn{2}{c|}{Market} & \multicolumn{2}{c}{DukeMTMC} \\
\multicolumn{1}{c|}{} &  & Rank1 & mAP & Rank1 & mAP \\ \hline
\multirow{3}{*}{Mask-Guided} & MaskGuided \cite{MaskReID} & 83.8 & 74.3 & - & - \\
 & MaskReID \cite{MaskReID} & 90.0 & 75.3 & 78.8 & 61.9 \\
 & SPReID \cite{SPReID} & 92.5 & 81.3 & 84.4 & 71.0 \\ \hline
\multirow{4}{*}{Attention-Based} & DuATM \cite{DuATM} & 91.2 & 75.7 & 80.5 & 63.8 \\
 & HA-CNN \cite{HA-CNN} & 91.4 & 76.6 & 81.2 & 62.3 \\
 & Mancs \cite{Mancs} & 93.1 & 82.3 & 84.9 & 71.8 \\ 
  & AANet \cite{AANet} & 93.9 & 83.4 & 84.9 & 71.8 \\ \hline
\multirow{2}{*}{GAN-Based} & Camstyle \cite{CamStyle} & 88.1 & 68.7 & 75.3 & 53.5 \\
 & PN-GAN \cite{PN-GAN} & 89.4 & 72.6 & 73.6 & 53.2 \\ \hline
\multirow{3}{*}{Global Feature} 
 & BagTricks \cite{StrongBaseline} & {94.5} & {85.9} & {86.4} & {76.4} \\
 & \textbf{BagTricks+APRA} & {\textbf{94.7}} & \textbf{86.9} & \textbf{87.8} & \textbf{78.2} \\
 & RGA \cite{RGA} & {95.0} & {88.4} & {87.8} & {78.4} \\
 & \textbf{RGA+APRA} & {\textbf{95.4}} & \textbf{88.7} & \textbf{88.8} & \textbf{78.6} \\ \hline 
\end{tabular}}
\end{center}
\vspace{-1em}
\caption{Comparison of state-of-the-art methods.}
\vspace{-0.5em}
\label{tab:SOTA}
\end{table}
\vspace{-1em}

\section{Conclusions}
Imbalanced performance among cameras remains an unexplored area in Person ReID. To study this, we formulated new camera-specific evaluation metrics and quantified the imbalance in popular benchmark datasets. To support our discovery, we contributed one of few privacy-aware Person ReID datasets that spans over 38 real-world surveillance cameras with natural imbalance. We designed the Adversarial Pairwise Reverse Attention (APRA) module, a plug-and-play component that learns high quality camera-invariant features to improve overall and bottleneck camera scores. Other factors like imbalance of training samples or pose variation across cameras could be interesting topics for further investigation. We hope that our study encourages exploration into this nascent space as Person ReID advances into practical application.

\section{Acknowledgements}
This work was supported by the Defence Science and Technology Agency (DSTA). It was carried out at the Rapid-Rich Object Search (ROSE) Lab at the Nanyang Technological University, Singapore.
\bibliographystyle{IEEEbib}
\bibliography{references-short}

\begin{thebibliography}{10}

\bibitem{Market-1501}
Zheng et~al,
\newblock ``{Scalable Person Re-identification: A Benchmark},''
\newblock in {\em ICCV}, 2015.

\bibitem{DukeMTMC-reID}
Zheng et~al,
\newblock ``{Unlabeled Samples Generated by GAN Improve the Person
  Re-identification Baseline in Vitro},''
\newblock in {\em ICCV}, 2017.

\bibitem{CUHK03}
Wang et~al,
\newblock ``{Joint Learning of Single-Image and Cross-Image Representations for
  Person Re-identification},''
\newblock in {\em CVPR}, 2016.

\bibitem{Huang2019Multi-PseudoRe-Identification}
Huang et~al,
\newblock ``{Multi-Pseudo Regularized Label for Generated Data in Person
  Re-Identification},''
\newblock {\em TIP}, 2019.

\bibitem{Lin2017End-to-EndRe-Identification}
Lin et~al,
\newblock ``{End-to-End Correspondence and Relationship Learning of Mid-Level
  Deep Features for Person Re-Identification},''
\newblock in {\em DICTA}, 2017.

\bibitem{TriNet}
Hermans et~al,
\newblock ``{In Defense of the Triplet Loss for Person Re-Identification},''
\newblock in {\em arXiv preprint}, 2017.

\bibitem{Quadruplet}
Chen et~al,
\newblock ``{Beyond Triplet Loss: A Deep Quadruplet Network for Person
  Re-identification},''
\newblock in {\em CVPR}, 2017.

\bibitem{IDE}
Zheng et~al,
\newblock ``{Person Re-identification: Past, Present and Future},''
\newblock {\em arXiv preprint}, 2016.

\bibitem{ImageNet}
Deng et~al,
\newblock ``{ImageNet: A large-scale hierarchical image database},''
\newblock in {\em CVPR}, 2009.

\bibitem{Sun2017SVDNetRetrieval}
Sun et~al,
\newblock ``{SVDNet for Pedestrian Retrieval},''
\newblock in {\em ICCV}, 2017.

\bibitem{Ensemble}
Paisitkriangkrai et~al,
\newblock ``{Learning to rank in person re-identification with metric
  ensembles},''
\newblock in {\em CVPR}, 2015.

\bibitem{MGN}
Wang et~al,
\newblock ``{Learning Discriminative Features with Multiple Granularities for
  Person Re-Identification},''
\newblock in {\em ACM MM}, 2018.

\bibitem{StrongBaseline}
Luo et~al,
\newblock ``{Bag of Tricks and a Strong Baseline for Deep Person
  Re-Identification},''
\newblock in {\em CVPRW}, 2019.

\bibitem{HA-CNN}
Li~et~al,
\newblock ``{Harmonious Attention Network for Person Re-identification},''
\newblock in {\em CVPR}, 2018.

\bibitem{AANet}
Tay et~al,
\newblock ``{AANet: Attribute Attention Network for Person
  Re-Identifications},''
\newblock in {\em CVPR}, 2019.

\bibitem{DuATM}
Si~et~al,
\newblock ``{Dual Attention Matching Network for Context-Aware Feature Sequence
  Based Person Re-identification},''
\newblock in {\em CVPR}, 2018.

\bibitem{CamStyle}
Zhong et~al,
\newblock ``{Camera Style Adaptation for Person Re-identification},''
\newblock in {\em CVPR}, 2018.

\bibitem{CycleGAN}
Zhu et~al,
\newblock ``{Unpaired Image-to-Image Translation Using Cycle-Consistent
  Adversarial Networks},''
\newblock {\em ICCV}, 2017.

\bibitem{SingleCamera}
Zhang et~al,
\newblock ``{Single Camera Training for Person Re-Identification},''
\newblock in {\em AAAI}, 2020.

\bibitem{Zhu2019Intra-cameraBenchmark}
Zhu et~al,
\newblock ``{Intra-camera supervised person re-identification: A new
  benchmark},''
\newblock in {\em ICCVW}, 2019.

\bibitem{MSMT17}
Wei et~al,
\newblock ``{Person Transfer GAN to Bridge Domain Gap for Person
  Re-identification},''
\newblock in {\em CVPR}, 2018.

\bibitem{CBAM}
Woo et~al,
\newblock ``{CBAM: Convolutional block attention module},''
\newblock in {\em ECCV}, 2018.

\bibitem{DANN}
Ganin et~al,
\newblock ``{Domain-Adversarial Training of Neural Networks},''
\newblock {\em JMLR}, 2016.

\bibitem{MaskGuided}
Song et~al,
\newblock ``{Mask-Guided Contrastive Attention Model for Person
  Re-identification},''
\newblock in {\em CVPR}, 2018.

\bibitem{MaskReID}
Qi~et~al,
\newblock ``{A Mask Based Deep Ranking Neural Network for Person Retrieval},''
\newblock in {\em ICME}, 2019.

\bibitem{SPReID}
Kalayeh et~al,
\newblock ``{Human Semantic Parsing for Person Re-identification},''
\newblock in {\em CVPR}, 2018.

\bibitem{Mancs}
Wang et~al,
\newblock ``{Mancs: A Multi-task Attentional Network with Curriculum Sampling
  for Person Re-Identification},''
\newblock in {\em ECCV}, 2018.

\bibitem{PN-GAN}
Qian et~al,
\newblock ``{Pose-normalized image generation for person re-identification},''
\newblock in {\em ECCV}, 2018.

\bibitem{RGA}
Zhizheng Zhang, Cuiling Lan, Wenjun Zeng, Xin Jin, and Zhibo Chen,
\newblock ``Relation-aware global attention for person re-identification,''
\newblock {\em CVPR}, 2020.

\end{thebibliography}

\end{document}